\documentclass{article}

\PassOptionsToPackage{numbers, compress}{natbib}

\usepackage[final]{neurips_2022_ml4ps}

\usepackage[utf8]{inputenc} 
\usepackage[T1]{fontenc}    
\usepackage{hyperref}       
\usepackage{url}            
\usepackage{booktabs}       
\usepackage{amsfonts}       
\usepackage{nicefrac}       
\usepackage{microtype}      
\usepackage{xcolor}         
\usepackage{amsmath}
\usepackage{enumitem}

\usepackage{bm}
\usepackage{amsmath}
\usepackage{graphicx}
\usepackage{sidecap}

\title{Validation Diagnostics for SBI algorithms based on Normalizing Flows}

\author{%
  Julia Linhart
  \\
  Université Paris-Saclay, Inria, CEA\\
  Palaiseau 91120, France \\
  \texttt{julia.linhart@inria.fr} \\
  \And
    Alexandre Gramfort
  \\
  Université Paris-Saclay, Inria, CEA\\
  Palaiseau 91120, France \\
  \texttt{alexandre.gramfort@inria.fr} \\  
  \AND
    Pedro L. C. Rodrigues\\
    Univ. Grenoble Alpes, Inria, CNRS, Grenoble INP, LJK \\
    Grenoble 38000, France\\
  \texttt{pedro.rodrigues@inria.fr} \\
}

\begin{document}

\maketitle

\begin{abstract}
Building on the recent trend of new deep generative models  known as Normalizing Flows (NF), simulation-based inference (SBI) algorithms can now efficiently accommodate arbitrary complex and high-dimensional data distributions. The development of appropriate validation methods however has fallen behind. 
Indeed, most of the existing metrics either require access to the true posterior distribution, or fail to provide \textit{theoretical guarantees} on the consistency of the inferred approximation beyond the one-dimensional setting. This work proposes easy to interpret validation diagnostics for multi-dimensional conditional (posterior) density estimators based on NF. It also offers theoretical guarantees based on results of local consistency. The proposed workflow can be used to check, analyse and guarantee consistent behavior of the estimator. The method is illustrated with a challenging example that involves tightly coupled parameters in the context of computational neuroscience. This work should help the design of better specified models or drive the development of novel SBI-algorithms, hence allowing to build up trust on their ability to address important questions in experimental science.
\end{abstract}

\section{Introduction}

Recent advances in computing have led to a new generation of expressive simulators used to study complex systems in many scientific fields~\citep{Cranmer2020}. They implicitly encode the intractable likelihood $p(x\mid\bm{\theta})$ of the underlying mechanistic model which relates observed data $x \in \mathbb{R}^d$ to scientifically meaningful internal parameters $\bm{\theta} \in \mathbb{R}^m$.
To perform statistical inference in this setting, one can recur to simulation based inference (SBI)~\citep{Cranmer2020} 
to approximate the posterior distribution $p(\bm{\theta} \mid x)$ using samples from the joint pdf $p(x, \bm{\theta})$. In this work, we consider SBI methods based on normalizing flows~\citep{Papamakarios2021}, which are invertible neural networks that can be trained via maximum likelihood. Once the flow is trained, and for any new observation $x$, one can directly evaluate the estimated density over the parameter space (i.e. $\bm{\theta}$-space)~\citep{Goncalves2020}, draw samples to construct confidence regions~\citep{Masserano2022}, etc.

Flow-based SBI has been used in many recent applied works~\citep{Lueckman2021, Bittner2021, Goncalves2020}, but it lacks an important feature before becoming a technology for experimental science: \emph{validation}. Ideally, one would like to have a method that provides finite-sample guarantees of nominal coverage (or calibration) of the estimated posterior regions, but also ensures that the approximation fits the true underlying posterior of the model when
new data is observed. While solutions have been proposed to provide such finite-sample guarantees~\citep{Dalmasso2021, Masserano2022}, assessing the convergence and consistency of the underlying inference remains a challenging task \citep{Lueckman2021}. Simulation based calibration (SBC)~\citep{Talts2018} is arguably the most popular metric for validating posterior approximations in the applied SBI community~\citep{Boelts2022}, but it only provides necessary conditions for consistency and fails to give any insight on the \textit{local} behavior of the estimator. Moreover, it is a univariate procedure, thus ignoring any information about the coupling between parameters. \citet{Zhao2021} recently proposed local coverage tests (LCT).
They leverage machine learning to 
evaluate their test quantities
on different locations of the feature space (i.e. $x$-space). In 1D, the probability integral transform (PIT) provides \emph{necessary and sufficient} conditions for consistency, which is not the case for the proposed multivariate extensions (e.g. HPD).



In this work, we present a multivariate version of LCT for density models based on normalizing flows. Our method comes with the same theoretical guarantees on local consistency and interpretable diagnostics as provided by \citep{Zhao2021} in 1D.  
We also present a workflow that can be used as a practical user guide.
Lastly, we provide numerical illustrations on a well known model from computational neuroscience~\citep{Jansen1995}, and demonstrate the importance of trustworthy diagnostics for multivariate posterior distributions when correlations between parameter variables can play an important role.

\section{Methods}\label{sec-methods}
Our conditional density estimator $q_\phi$ is a normalizing flow defined for samples $\bm{\theta} \in \mathbb{R}^m$ and is conditioned on observations $x \in \mathbb{R}^d$. It uses a Gaussian base distribution $p(\bm{z}) = \mathcal{N}(0, \bm{I}_m)$ and a bijective transformation defined for every $x$, $T_{\phi}(.;x) :=  (T_{\phi,1}(.;x), \dots, T_{\phi,m}(.;x))$,
with Jacobian $J_{T_{\phi}}(.; x)$,  
such that $q_{\phi}(\bm{\theta}\mid x) = p(T_{\phi}(\bm{z};x)) = p(\bm{z})\left|\operatorname{det} J_{T_{\phi}}(\bm{z}; x)\right|^{-1}~$.


Our goal is to evaluate the \emph{local consistency}~\citep{Kim2018} of $q_{\phi}$ with respect to the true posterior density, i.e. whether for a given $x$ the following null hypothesis holds:
\begin{equation}
\label{eq:null_hypothesis}
    \mathcal{H}_0(x): q_{\phi}(\bm{\theta} \mid x) = p(\bm{\theta} \mid x), \quad \forall \bm{\theta} \in \mathbb{R}^m~.
\end{equation}

We define the multivariate \textit{probability integral transform} (PIT) of $\bm{\theta}$ at $x$ and associated to $q_{\phi}$ as the vector of $m$ one-dimensional projections:
\begin{equation}\label{multiPIT}
    \text{PIT}_m(\bm{\theta},x,q_{\phi}) = [P_1(\bm{\theta},x),\dots, P_m(\bm{\theta},x)], \quad P_i(.,x) = F_{\mathcal{N}(0,1)}\circ T_{\phi,i}^{-1}(.;x), ~\forall i \in [1,m]
\end{equation}
where $F_{\mathcal{N}(0,1)}$ is the cumulative distribution function (c.d.f.) of the univariate normal distribution. 

\textbf{Theorem 1: Local Consistency and multivariate PIT.}
\textit{For any $x \in \mathbb{R}^d$, the null hypothesis $\mathcal{H}_0(x)$ holds if, and only if, the covariates of $\text{PIT}_m(\bm{\theta}, x, q_{\phi})$ conditioned on $x$ are mutually independent and uniformly distributed over $(0,1)$}. We refer to Appendix~\ref{sec:appendix} for a detailed proof of this result. 
\begin{align*}
    p(\bm{\theta} \mid x) = q_{\phi}(\bm{\theta} \mid x)  
    & \iff p(T^{-1}_{\phi}(\bm{\theta},x) \mid x) = p(\bm{z})\\
    & \iff  p(T^{-1}_{\phi,1}(\bm{\theta},x), \dots T^{-1}_{\phi,m}(\bm{\theta},x)\mid x) = \mathcal{N}(0,\bm{I}_m)\\ 
    & \iff \left\{
    \begin{array}{ll}
    p(P_i(\bm{\bm{\theta}},x)\mid x) = \mathcal{U}(0,1) \quad \forall i \in [1,m] \quad \text{and} \\[0.5em]
     \{P_i(\bm{\theta},x) \mid x\}_{i = 1, \dots, m} \quad \text{are mutually independent}
     \end{array} \right.
\end{align*}

We can now verify the null hypothesis of local consistency $\mathcal{H}_0(x)$ (\ref{eq:null_hypothesis}) via $m$ statistical tests for the uniformity of the 1D local PIT covariates and an additional test for their mutual independence.

\textbf{Multivariate Local Coverage Tests (LCT).} 
Noting that for every $ i = 1, \dots, m$ we have 
\begin{equation}
\label{eq:uniformity-test}
P_i(\bm{\theta},x) \mid x \sim \mathcal{U}(0, 1) \iff \forall \alpha \in [0, 1]\quad r_{i,\alpha}(x) = \mathbb{P}(P_i(\bm{\theta},x) \leq \alpha \mid x) = \alpha   
\end{equation}
and following the same approach as in~\citep{Zhao2021}, we propose $m$ test statistics
\begin{equation}
\label{eq:test-statistic}
\mathcal{T}_{i}(x):=\frac{1}{|G|} \sum_{\alpha \in G}\Big(\widehat{r}_{i,\alpha}(x)-\alpha\Big)^2 \quad \forall i \in [1,m]~,
\end{equation}
where $G$ is a grid of $\alpha$-values and the estimators $\widehat{r}_{i, \alpha}$ are obtained by regressing $\mathbb{I}_{\{P_i(\bm{\theta},x) \leq \alpha\}}$ on $x$, which is optimal for $\mathbb{E}[\mathbb{I}_{\{P_i(\bm{\theta},x) \leq \alpha\}} \mid x] = r_{i, \alpha}(x)$ (when using appropriate loss-functions~\citep{Miller1993}). 
We refer to Algorithm 1 (resp. 2) in \citep{Zhao2021}
for computing the $p$-values (resp. confidence bands) associated to each test. Since there are $m$ independent tests, we recur to a Bonferroni correction of the $p$-values (resp. confidence levels)~\citep{Bland1995}. These tests come with interpretable graphical diagnostics such as PP-plots or histograms that depict distributional deviations in 1D, as shown in Figure \ref{fig:global}. 
If any of the uniformity tests is rejected, we reject $\mathcal{H}_0(x)$. If not, we proceed to the mutual independence test. 

We now assume that the covariates of $\text{PIT}_m$ conditioned on $x$ are uniformly distributed over $(0,1)$, which also means that every coordinate of the flow-transformation $T_{\phi}^{-1}(\bm{\theta}, x)$ follows a normal distribution \textit{given} $x$ (cf. Theorem 1). Their mutual independence is thus characterized by a covariance matrix
equal to the identity $\mathbf{I}_m$. We are currently working on how to perform this check in practice.

\textbf{The workflow in practice.} Let $\mathcal{D} = \{x_n, \bm{\theta}_n\}_{n=1}^N$ be a calibration dataset with $(x_n, \bm{\theta}_n) \sim p(x, \bm{\theta})$ which were \textit{not} used to train $q_\phi$. We use $\mathcal{D}$ to calculate the PIT values $\text{PIT}_m(\bm{\theta}_n, x_n, q_{\phi})$ (\ref{multiPIT}) and estimate our local test quantities as functions of $x$. We investigate the \emph{consistency} of $q_\phi$ in two parts:
\begin{enumerate}[leftmargin=2em, nosep]
    \item[(1)] \emph{Global consistency check}: first, we look at the global PIT-distribution, i.e. on average over the entire $x$-space. More specifically, we directly compute the empirical approximation of  $$\textstyle r_{i,\alpha}=\mathbb{P}(P_i(\bm{\theta}, x) \leq \alpha) \approx  \tfrac{1}{N}\sum_{n=1}^N\mathbb{I}_{\{P_i(\bm{\theta}_n, x_n)\leq \alpha \}}, \quad \forall i \in [1,m]$$ 
    with samples from $\mathcal{D}$ to test the global uniformity of each PIT-covariate 
    and check which one(s) might be responsible for making $q_\phi$ deviate from the true posterior distribution. Note that passing such global test is only a neccessary condition for consistency of $q_\phi$, as it is insensitive to covariate transformations in $x$-space \citep{Zhao2021} (and ignores the condition on mutual independence).
    \item[(2)] \emph{Local consistency check}: we construct $m \times |G|$ transformed datasets from $\mathcal{D}$
    $$\mathcal{D}_{i,\alpha} = \{(x_n, W_n^{i,\alpha})\}_{n = 1}^N , \quad \alpha \in G, \quad  \forall i \in [1,m]$$ where $W_n^{i,\alpha} = \mathbb{I}_{\{P_i(\bm{\theta}_n,x_n) \leq \alpha\}}$. We can then compute the test statistics defined in~\eqref{eq:test-statistic} for any new observation $x$ and check whether $\mathcal{H}_0(x)$ should be rejected or not. In the latter case we proceed to the mutual independence test. Only then can we conclude
    on the validity of $\mathcal{H}_0(x)$ according to Theorem 1. If the check in (1) passes, this allows us to \emph{guarantee} (or reject) consistency anywhere in $x$-space.
    Even if the check in (1) does not pass, analyzing local consistency allows to `open the box' and better understand why and where in $x$-space the estimator fails.
    In such situations -- as the goal is \textit{not to guarantee} local consistency -- it can be enough to test for the covariate-wise uniformity of $\text{PIT}_m$, putting the test for mutual independence aside. This approach was adopted for our numerical illustrations in Section \ref{sec-results}.
\end{enumerate}

\section{Numerical illustrations}\label{sec-results}
\begin{figure}
    \centering
    \includegraphics[width=\textwidth]{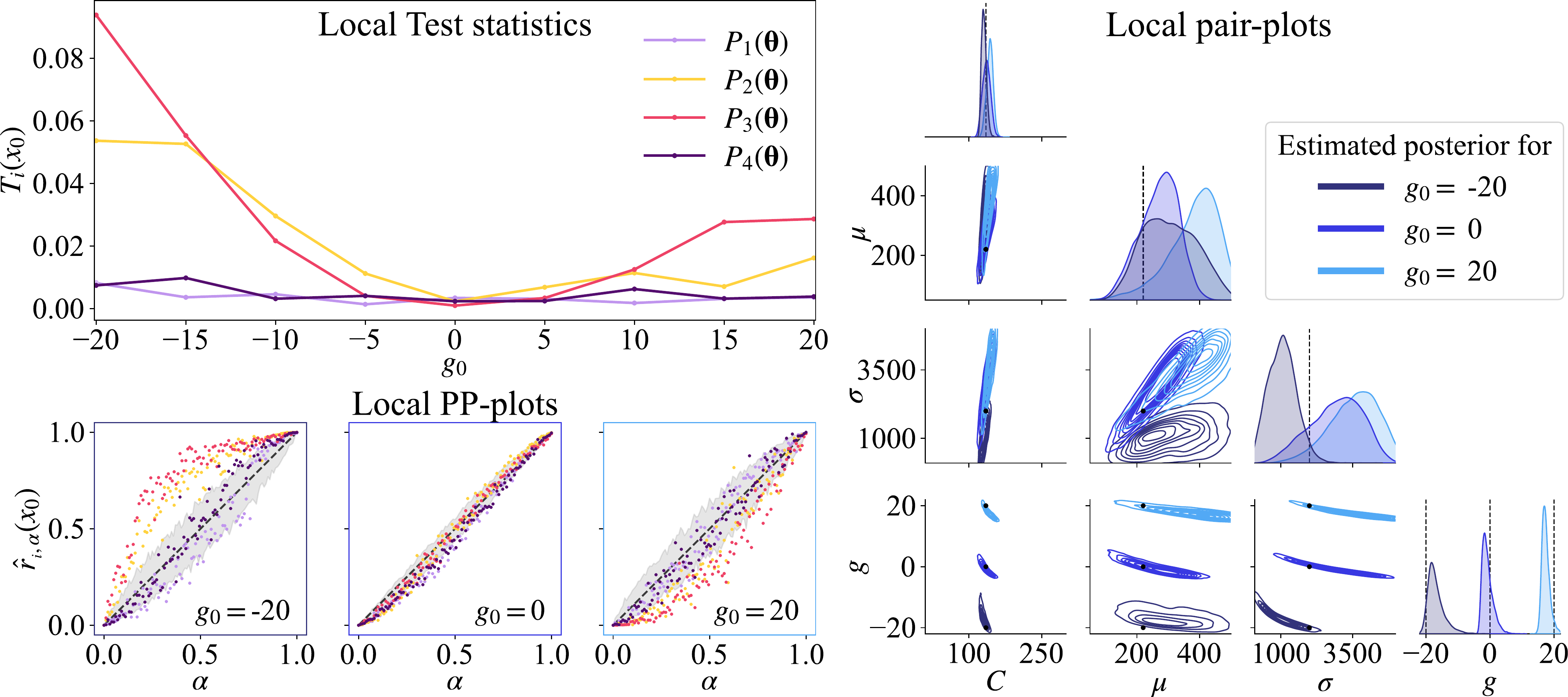}
    \caption{Local Consistency Analysis. (Left) Diagnostics obtained for every PIT-covariate, when evaluated for different $x_0$, simulated via the JRNMM with fixed parameters $(C_0, \mu_0, \sigma_0)$ and variable $g_0$ in $[-20,20]$. The test-statistics are plotted as a function of $g_0$ and indicate inconsistent behavior (values are not constantly close to zero). Below, the PP-plots report the nature (bias/dispersion) of these inconsistencies for $g_0 \in \{-20, 0, 20\}$. The gray zone indicates the $95\%$-confidence region of acceptance outside of which the uniformity test is rejected. (Right) Pair-plots of $q_{\phi}(\bm{\theta}\mid x_0)$ with ground-truth parameters $\bm{\theta}_0=(C_0, \mu_0, \sigma_0, g_0)$ (black dots and dashed lines) for $g_0 \in \{-20, 0, 20\}$.
}
    \label{fig:local}
\end{figure}

We apply our method to check the validity of a posterior estimate $q_\phi$ of the Jansen and Rit neural mass model (JRNMM)~\citep{Jansen1995}. This well known model from computational neuroscience takes parameters $\bm{\theta}=(C,\mu,\sigma,g) \in \mathbb{R}^4$ as input and generates time series $x\in \mathbb{R}^{1024}$ with properties similar to brain signals obtained in neurophysiology. Parameter $C$ influences the oscillatory behavior of the signals and $(\mu, \sigma)$ characterize their amplitude. The gain factor $g$ rescales the signals and models the effects of the amplifier used for measuring them in practice. Note that the coupling-effect of $g$ and $(\mu, \sigma)$ on the amplitude leads to intrinsic indeterminacies in the inversion of the model~\citep{Rodrigues2021}. Our approximation $q_\phi$ is a conditioned masked autoregressive flow (MAF)~\cite{Papamakarios2017} with 10 layers and implemented in the \texttt{sbi} package~\citep{Tejero2020}. Based on the setup in~\citep{Rodrigues2021}, we train $q_\phi$ on 50\,000 simulations from the JRNMM with a uniform prior defined on physiologically relevant values (cf. pair-plots in Figure \ref{fig:local}).

\begin{SCfigure}
    \centering
    \includegraphics[width=0.52\textwidth]{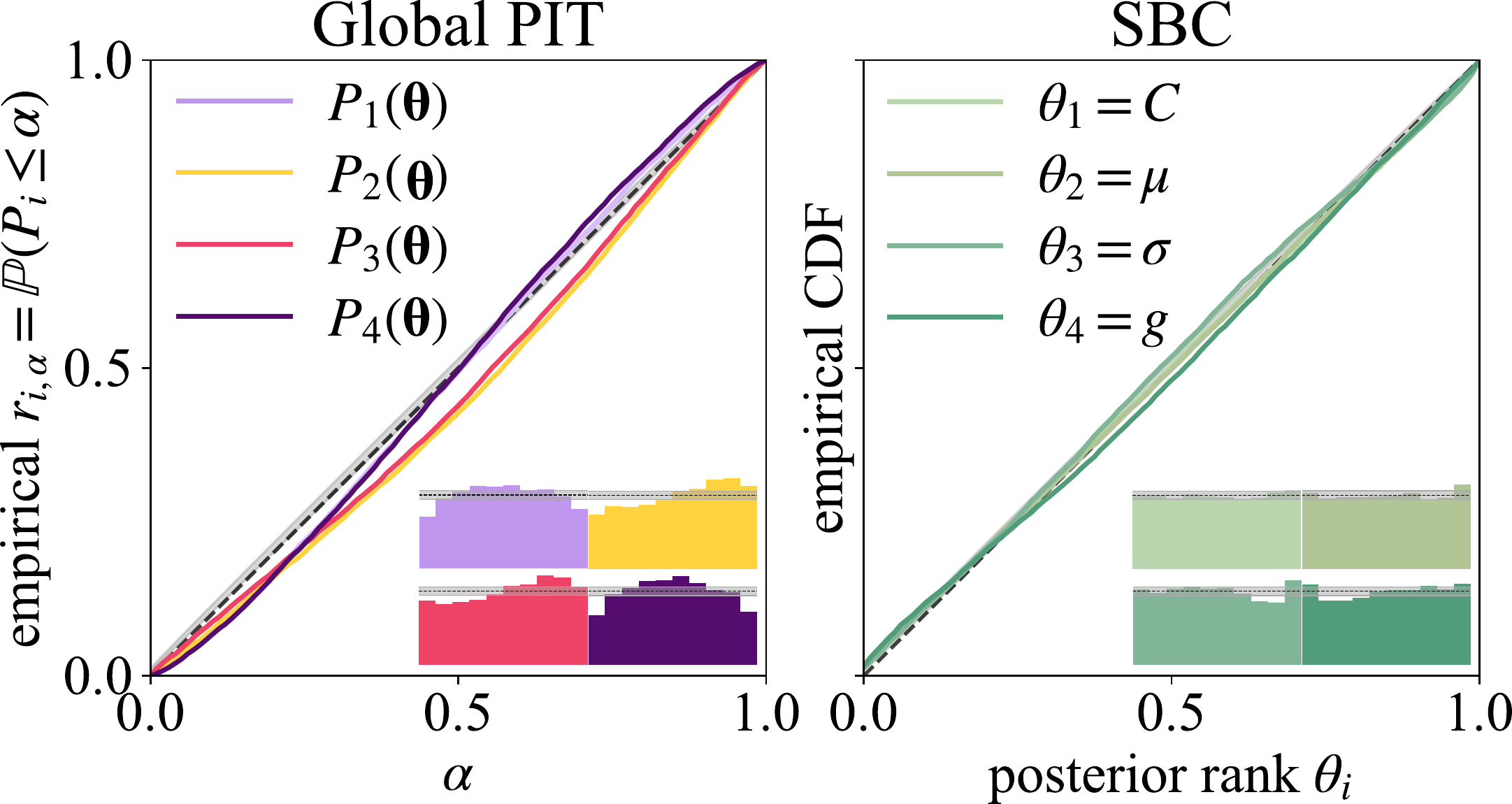}
    \caption{Global Consistency of the JRNMM posterior estimator: PP-plots and corresponding histograms obtained for every global PIT covariate $P_i(\bm{\theta})$ (right) and for SBC applied to every element $\theta_i$ of $\bm{\theta}$ (left). The gray zone indicates the $95\%$-confidence region of acceptance outside of which the uniformity test is rejected. SBC fails to reject the null hypothesis of global consistency.
}
    \label{fig:global}
\end{SCfigure}

\textbf{Results and discussion.} All test quantities are computed with a calibration dataset $\mathcal{D}$ containing 10\,000 simulations and a grid of $|G|=100$ $\alpha$-values in [0, 1]. 
Following the workflow described in Section~\ref{sec-methods}, we first check the global consistency of $q_\phi$. In the left part of Figure~\ref{fig:global}, our graphical diagnostics illustrate how the c.d.f. for every PIT-covariate deviates from the identity function (black dashed line), outside of the $95\%$-confidence region (in gray), thus rejecting the null hypothesis of global consistency. 
We compare our method to SBC (right part of Figure \ref{fig:global}) as implemented in the \texttt{sbi} package. We observe that SBC is unable to detect any inconsistencies in $q_\phi$.

We proceed to investigate the local consistency of $q_\phi$ on different locations of $x$-space. Note that since the global test for consistency has not passed, we may focus on the uniformity of the local PIT-covariates (cf. (2) of the workflow in Section \ref{sec-methods}). 
We consider a 1D subspace in $\bm{\theta}$-space, where $(C_0, \mu_0, \sigma_0)$ are fixed and  the gain $g_0$ varies in $[-20,20]$ to generate observations $x_0$. The upper-left part of Figure~\ref{fig:local} shows how the test statistics evolve with $g_0$. We observe a strong deviation from uniformity for covariates $P_2$ and $P_3$: they vary smoothly in a `U-shape', with \textit{higher values} as $g_0$ deviates from zero. We also generate local PP-plots (lower-left part of Figure~\ref{fig:local}) and observe positive (resp. negative) bias for small (resp. high) values of $g_0$. Our procedure shows that there are certain locations in $x$-space (here $g_0=0$) where $q_\phi$ performs well (i.e. test statistics are close to zero and PP-plots show little deviations from the $95\%$-confidence region in gray),
even though global consistency does not hold. These observations can be compared to the pair-plots (right part of Figure \ref{fig:local}) which shows the estimated posterior density over the entire $\bm{\theta}$-space.
We observe that gain values at the boundary of the prior support ($g_0=-20$ and $g_0=20$) induce discrepancies (cf. marginals plotted in the diagonals of the pair-plots in Figure \ref{fig:local}) that are detected by our PIT-based diagnostics. 


The above results 
are obtained by directly applying the method from \citep{Zhao2021} using their default regressor, \texttt{MLPClassifier} from scikit-learn~\citep{scikit-learn}. Indeed, this model is well suited for the binary target variables $W^{i,\alpha}$ and, based on neural networks, is able to scale to high-dimensional $x$-spaces. Furthermore, it has been shown~\citep{Miller1993} that for a large enough dataset $\mathcal{D}$ (and number of training iterations), the cross-entropy loss converges to the optimal solution of our regression problem. 
Note that although such method allows to capture and visualize local discrepancies,
it does not cope well with our high-dimensional, complex data-distributions (high variances represented by large confidence regions computed over $100$ trials).
Also, it requires the training of a large number ($m \times |G|$) of regressors (one for each $\bm{\theta}$-dimension and $\alpha$-value).
Finally, this method is only applicable for the uniformity tests. 
We are currently investigating algorithms that are computationally more efficient and more accurate in estimating the \emph{full} multivariate  local PIT-distribution. 

\textbf{Conclusions.} Numerical illustrations demonstrate that our diagnostics capture well the inconsistencies in $q_\phi$ with respect to the true JRNMM posterior. To be precise, the proposed validation method is statistically more powerful and computationally more efficient than SBC and, importantly, it allows for a local analysis that reveals where in $x$-space the estimations should be improved. 
Our method exploits useful properties of modern normalizing flows. Indeed, contrarily to other SBI-strategies~\citep{Hermans2020}, they allow for  efficient density evaluation and use bijective transformations involving the mixing between elements of $\bm{\theta}$. These transformations can be interpreted as a set of 1D-projections of $\bm{\theta}$, including information about its joint p.d.f. (i.e. interactions between its elements $\theta_i$). The covariates of our multivariate PIT therefore define fast and easy to compute 1D test-quantities on which we can perform univariate LCTs that do not completely ignore correlations in the $\bm{\theta}$-space.  
Combined with a test to check their mutual independence, this would provide theoretical guarantees for consistency, which is not the case for HPD, the existing multivariate version of LCT proposed in \citep{Zhao2021}.

Ongoing work will investigate the link between PIT-covariates and their deviations from uniformity with the actual parameters of the model. This could reveal the true nature of their coupling which can be used for the development of better specified models (e.g. hierarchical posterior estimation~\citep{Rodrigues2021, Rouillard2022}). Ultimately, the goal is to use our diagnostics as a tool for model selection.



\section*{Broader Impact}
This work tackles an important open question in simulation based inference. We introduce theoretically valid and interpretable validation diagnostics that scale to both high-dimensional data and parameter spaces. Our contribution should help to further improve SBI methods and drive the design of better specified models, hence allowing to build up trust on their ability to address important questions in experimental science. Moreover, our work is in line with many other tentatives in the machine learning community of ensuring the quality and calibration of complex models based on neural networks.

\section*{Acknowledgments}

Julia Linhart is recipient of the Pierre-Aguilar Scholarship and thankful
for the funding of the Capital Fund Management (CFM). Alexandre Gramfort thanks the support of the ANR BrAIN (ANR-20-CHIA0016) grant.

\bibliography{valDiags}

\newpage
\appendix

\section{Appendix}
\label{sec:appendix}
\textbf{Proof of Theorem 1.} Let $x \in \mathbb{R}^d$. We consider the random variables defined on $\mathbb{R}^m$: $\Theta \sim p(\boldsymbol{\theta} \mid x)$ and $\Theta_\phi \sim q_\phi(\boldsymbol{\theta}\mid x)$.
Our local null hypothesis (cf. Equation~\ref{eq:null_hypothesis}), can be rewritten as
\begin{equation}\label{null-hyp-proof}
\mathcal{H}_0(x)~:~\text{Prob}\Big(\Theta \in {\Omega_\theta}\Big) = \text{Prob}\Big(\Theta_\phi \in {\Omega_\theta}\Big)  \quad \forall \Omega_{\theta} \subset \mathbb{R}^m
\end{equation}

From the definition of our normalizing flow $q_{\phi}$ with bijective transformation $T_{\phi}$ and normal base distribution, there exists $Z\sim p(\mathbf{z})=\mathcal{N}(0,\mathbf{I}_m)$, such that $\Theta_{\phi} = T_{\phi}(Z,x)$. 

Replacing $\Theta_{\phi}$ by $T_{\phi}(Z,x)$ in (\ref{null-hyp-proof}) and defining $\Omega_z = T_{\phi}^{-1}(\Omega_\theta; x)$, we have
$$
\begin{array}{rcl}
\text{Prob}\Big(\Theta \in {\Omega_\theta}\Big) = \text{Prob}\Big(T_\phi(Z; x) \in {\Omega_\theta}\Big)  &\iff& \text{Prob}\Big(T_\phi^{-1}(\Theta; x) \in \Omega_z\Big) = \text{Prob}\Big(Z \in \Omega_z\Big)
\end{array}
$$
Considering $T^{-1}_{\phi, i}$, the $i$-th coordinate of $T^{-1}_\phi$, and $Z_1, \dots, Z_m$ the mutually independent and normally distributed  covariates of $Z \sim \mathcal{N}(0,\mathbf{I}_m)$, we can rewrite the right part of the above equivalence as:
\begin{equation}\label{factorization}
\text{Prob}\Big(T_{\phi, 1}^{-1}(\Theta; x) \in \Omega_{z,1} \cap \dots \cap T_{\phi, m}^{-1}(\Theta; x) \in \Omega_{z,m} \Big) = \displaystyle\prod_{i = 1}^m  \text{Prob}\Big(Z_i \in \Omega_{z,i}\Big)
\end{equation}
where the $\Omega_{z,i}$ are projections of the set $\Omega_z$ on each of its dimensions. 

Note that since the above equivalence is true for \textit{any} choice of $\Omega_{\theta}$, hence of $\Omega_{z}$, we can write what happens when we fix a given $\Omega_{z,1}$ and let $\Omega_{z,2} = \dots = \Omega_{z,m} = \mathbb{R}$:
$$
\text{Prob}\Big(T_{\phi, 1}^{-1}(\Theta; x) \in \Omega_{z,1}  \Big) = \text{Prob}\Big(Z_1 \in \Omega_{z,1}\Big)
$$
Doing the same for all other coordinates and using these equalities in (\ref{factorization}), we end up with
\begin{align}\label{norm}
\forall i \in [1,m], \quad \forall \Omega_{z,i} \subset \mathbb{R}, \quad \text{Prob}\Big(T_{\phi, i}^{-1}(\Theta; x) \in \Omega_{z,i}  \Big) & = \text{Prob}\Big(Z_i \in \Omega_{z,i}\Big)
\\ \label{mut_ind}
\text{and} \quad \text{Prob}\Big(T_{\phi, 1}^{-1}(\Theta; x) \in \Omega_{z,1} \cap \dots \cap T_{\phi, m}^{-1}(\Theta; x) \in \Omega_{z,p} \Big) & = \displaystyle\prod_{i = 1}^m  \text{Prob}\Big(T_{\phi, i}^{-1}(\Theta; x) \in \Omega_{z,i}  \Big)
\end{align}
where (\ref{mut_ind}) is the definition of mutual independence itself.

We now apply the c.d.f.  $F_{\mathcal{N}(0,1)}$ to each random variable in (\ref{norm}) and (\ref{mut_ind}). Mutual independence stays true and the probability integral transform theorem (in 1D) states that $F_{\mathcal{N}(0,1)}(Z_i) \sim \mathcal{U}(0,1)$. We therefore get that the null hypothesis $\mathcal{H}_0(x)$ holds if, and only if,
\begin{equation}\label{conclusion}
\left\{
\begin{array}{ll}
 P_i(\Theta, x) = F_{\mathcal{N}(0,1)}(T_{\phi, i}^{-1}(\Theta; x)) \sim \mathcal{U}(0,1), \quad \forall i = [1,m] \quad \textit{and} \\[0.5em]
\{P_i(\Theta,x) \}_{i = 1, \dots, m} \quad \textit{are mutually independent}
\end{array}\right.
\end{equation}

The result in Theorem 1 directly follows from rewriting (\ref{null-hyp-proof}) and (\ref{conclusion}) with initial notations from Section \ref{sec-methods}: remember that $\Theta \sim p(\bm{\theta} \mid x)$, so  $P_i(\Theta,x) \sim p(P_i(\bm{\theta}, x) \mid x)$  and we get:
\begin{equation*}
    p(\bm{\theta} \mid x) = q_{\phi}(\bm{\theta} \mid x)  
    \iff   \left\{
    \begin{array}{ll}
    p(P_i(\bm{\bm{\theta}},x)\mid x) = \mathcal{U}(0,1) \quad \forall i \in [1,m] \quad \textit{and} \\[0.5em]
     \{P_i(\bm{\theta},x) \mid x\}_{i = 1, \dots, m} \quad \textit{are mutually independent}
     \end{array} \right.
\end{equation*}
where $P_i(\bm{\theta}, x)$ is the $i^{th}$ covariate of PIT$_m(\bm{\theta}, x, q_{\phi})$, the multivariate PIT of $\bm{\theta}$ at $x$, associated to $q_{\phi}$.

\textit{Conclusion:} The null hypothesis $\mathcal{H}_0(x)$ holds if, and only if, the covariates of PIT$_m$ conditioned on $x$ are mutually independent and uniformly distributed  over $(0,1)$.

\end{document}